\documentclass{article}
\usepackage{spconf,amsmath,graphicx}
\usepackage{epstopdf}
\usepackage{booktabs}
\usepackage{xcolor}
\usepackage{comment}
\usepackage{amsfonts}
\usepackage{subfigure}
\usepackage{hyperref}
\usepackage{balance}

\DeclareMathOperator*{\argmax}{arg\,max}
\DeclareMathOperator*{\argmin}{arg\,min}
\DeclareMathOperator{\tr}{tr}

\title{Multi-View Information Bottleneck without Variational Approximation}
\name{Qi Zhang$^1$, Shujian Yu$^{1,2}$\sthanks{To whom correspondence should be addressed (\texttt{yusj9011@gmail.com};  \texttt{chenbd@mail.xjtu.edu.cn}).}\sthanks{S. Yu was funded by the Research Council of Norway grant no. 309439.}, Jingmin Xin$^1$, Badong Chen$^{1*}$\sthanks{B. Chen was funded by the National Natural Science Foundation of China with grant numbers (62088102, U21A20485, 61976175).}
}
\address{$^1$Xi'an Jiaotong University, Xi'an 710049, Shaanxi, China\\
$^2$UiT - The Arctic University of Norway, 9037 Troms{\o}, Norway}

\begin{document}
\maketitle

\begin{abstract}
%Through fusion of complementary information between different views, multi-view learning is beneficial to improve the performance of classification task. In this paper, information bottleneck principle is used in supervised multi-view learning, which can effectively drop out the redundant information from each view and eliminate the noise information to some extent. Meanwhile, in stead of using variational approximation, the matrix-based R{\'e}nyi's $\alpha$-order entropy functional is used to directly optimize objective, which avoids uncertainty caused by variational approximation and hypothesis distribution. We conduct extensive experiments on synthetic datasets and real-world datasets, which proved that the method base on information bottleneck principle can effectively drop out redundant information contained in the different views. At the same time noise elimination experiment shows that our method has stronger robustness and have better generalization. Code of MEIB is available at~\url{https://github.com/archy666/MEIB}. 
By ``intelligently" fuse the complementary information across different views, multi-view learning is able to improve the performance of classification task. In this work, we extend the information bottleneck principle to supervised multi-view learning scenario and use the recently proposed matrix-based R{\'e}nyi's $\alpha$-order entropy functional to optimize the resulting objective directly, without the necessity of variational approximation or adversarial training. 
%We conduct experiments in both synthetic datasets and real-world datasets, which proved that the method base on information bottleneck principle can effectively drop out redundant information contained in the different views. At the same time noise elimination experiment shows that our method has stronger robustness and have better generalization. 
Empirical results in both synthetic and real-world datasets suggest that our method enjoys improved robustness to noise and redundant information in each view, especially given limited training samples. Code is available at~\url{https://github.com/archy666/MEIB}. 

\end{abstract}


\section{Introduction}
\label{sec:intro}
Multi-view learning becomes popular in many real-world applications, such as multi-omics data including genomics, epigenomics, proteomics, metabolomics~\cite{lee2021variational}. Traditional methods of multi-view learning include canonical correlation analysis (CCA)~\cite{chaudhuri2009multi} and its nonlinear extensions like kernel canonical correlation analysis (KCCA)~\cite{arora2012kernel}. However, both CCA and KCCA are restricted to only two views.

%which uses linear correlation coefficient to analyze correlation. By linear representation of data to reduce the dimensionality of high dimensional data. However, When linear representation of data cannot be achieved, kernel function idea can be used to map data to high dimensional space, and then the corresponding correlation coefficient and linear relationship can be obtained by CCA, so as to realize the nonlinear expansion of CCA, namely Kernel Canonical Correlation Analysis (KCCA)~\cite{arora2012kernel}. 

Due to the remarkable success achieved by deep neural networks (DNNs), there is a recent trend to leverage the power of DNNs to improve the performances of multi-view learning~\cite{yan2021deep}. Andrew \emph{et~al.}~\cite{andrew2013deep} proposed deep canonical correlation analysis (DCCA) to perform complex nonlinear transformation on multi-view data. Wang \emph{et~al.}~\cite{wang2015deep} proposed deep canonically correlated autoencoders (DCCAE) to further improve the performance of DCCA by adding an autoencoder reconstruction error regularization.

%To obtain minimum sufficient features, information bottleneck (IB) theory~\cite{tishby1999information} is extended to multi-view learning, but the mutual information terms in information bottleneck principle are computationally intractable. 

%By learning a minimum sufficient representation from training samples, 

Recently, the notable Information Bottleneck (IB) principle~\cite{tishby1999information} has been extended to multi-view learning problem to compress redundant or task-irrelevant information in the input views and only preserve the most task-relevant features~\cite{xu2014large,lou2013multi}. However, parameterizing the IB principle with DNNs is not a trivial task. A notorious reason is that the mutual information estimation in high-dimensional space is widely acknowledged as intractable or infeasible. To this end, existing deep multi-view IB approaches (e.g.,~\cite{lee2021variational, wang2019deep, aguerri2019distributed,federici2020learning,song2021multicolor}) usually use the variational approximation or adversarial training to maximize a lower bound of the original IB objective. However, the tightness of those derived lower bounds is hard to guarantee in practice, especially if there are only limited training data. 

%\cite{lee2021variational, wang2019deep, aguerri2019distributed, wan2021multi} \emph{et~al.}~ employed the variational inference method to optimize the original objective. The general process can be understood as finding a lower bound on the original objective first, and then finding the maximum value of this lower bound. This maximum is considered to be an approximation of the original objective minimum. Although it is common to use variational approximation with reparameterization trick~\cite{kingma2013auto} to estimate mutual information, the tightness of this approximation is difficult to guarantee.

% \commentSY{what do you mean by ``target"? it should be ``objective"? right? ... Are there only two papers [7,8] on deep multi-view learning? ... what about another TPAMI? [1] and many others ????}

%Compared with KCCA, DCCA has the advantage of parameter optimization without inner product. 
% autoencoder regularization terms for optimizing the canonical correlation between bottleneck features and autoencoder 

% In~\cite{hu2020dmib}, Hu \emph{et al.} proposed a dual-correlated multivariate information bottleneck method to explore both inter-feature correlations and inter-cluster correlations. An information maximization function is used to integrate the two correlations and boosts the overall clustering performance for multi-view learning. 

%In this work we employ matrix-based R{\'e}nyi's $\alpha$-order entropy functional to compute mutual information values instead of using variational approximation and hypothesis distribution to avoid variational boundary problems. 

In this work, instead of evaluating a lower bound of mutual information terms $I(X;Z)$ or $I(Y;Z)$ in deep multi-view information bottleneck, we demonstrate that it is feasible to directly optimize the IB objective without any variational approximation or adversarial training. We term our methodology \textbf{M}ulti-view matrix-based \textbf{E}ntropy \textbf{I}nformation \textbf{B}ottleneck (MEIB) and make the following contributions:
\begin{itemize}
   \item  We introduce the matrix-based R{\'e}nyi's $\alpha$-order entropy functional~\cite{giraldo2014measures} to estimate $I(X;Z)$ directly, which makes our MEIB enjoys a simple and tractable objective but also provides a deterministic compression of input variables from different views.
   \item Empirical results suggest that our MEIB outperforms other competing methods in terms of robustness to noise and redundant information, especially given limited training samples in each view.
    \item Our studies also raise a few new insights to the design and implementation of deep multi-view IB methods in the future, as will be discussed in the last section. 
\end{itemize}
    
\section{Preliminaries}
\label{sec:Preliminaries}
\subsection{Information Bottleneck Principle}
\label{ssec:IB Principle}

%Information bottleneck (IB) \cite{tishby1999information} is an information theory principle regards extract data features as a process of data compression. It is a trade-off between sufficiency and redundancy of information. Therefore, the features compressed from original data by IB principle will increase robustness for downstream tasks. Specifically, assuming that the original data is $X$, the task-related target is $Y$, and the feature extracted from $X$ is $Z$. Thus, IB principle can be formulated by 
%\begin{equation}
%     \max _{Z}{I(Y,Z)-\beta I(X,Z)},
%\end{equation}
%where $\beta$ is a trade-off factor to balance sufficiency $I(Y,Z)$ and redundancy $I(X,Z)$. 

Let us denote the input random variable as $X$ and desired output variable as $Y$ (e.g., class labels). The information bottleneck (IB) approach~\cite{tishby1999information} considers extracting a compressed representation $Z$ from $X$ that is relevant for predicting $Y$. Formally, this objective is formulated as finding $Z$ such that the mutual information $I(Y;Z)$ is maximized, while keeping mutual information $I(X;Z)$ below a threshold $\alpha$: 
\begin{equation}\label{eq:IB_original}\small
    Z^* = \argmax_{Z\in\Delta} I(Y;Z)\quad \text{s.t.} \quad I(X;Z)\leq \alpha,
\end{equation}
where $\Delta$ is the set of random variables $Z$ that obeys a Markov chain $Y - X - Z$, which means that $Z$ is conditional independent of $Y$ given $X$, and any information that $Z$ has about $Y$ is from $X$. 
%that obey the Markov chain $Y \leftrightarrow Z \leftrightarrow X$.  In deep variational IB, the Markov chain $Y \leftrightarrow Z \leftrightarrow X$ appears by construction  \cite{wieczorek2020difference}.

In practice, one is hard to solve the above constrained optimization problem of Eq.~(\ref{eq:IB_original}), and $Z$ is found by maximizing the following IB Lagrangian:
\begin{equation}\label{eq:IB_Lagrangian}\small
    \mathcal{L}_{IB}=I(Y;Z) - \beta I(X;Z),
\end{equation}
where $\beta>0$ is a Lagrange multiplier that controls the trade-off between the \textbf{sufficiency} (the performance on the task, as quantified by $I(Y;Z)$) and the \textbf{minimality} (the complexity of the representation, as measured by $I(X;Z)$). In this sense, the IB principle also provides a natural approximation of \emph{minimal sufficient statistic}~\cite{gilad2003information}.

\subsection{Information Bottleneck for Multi-view Learning} 
\label{ssec: Multi-view information bottleneck}
Multi-view learning comes down to the problem of machine learning from data represented by multiple distinct feature sets~\cite{sun2013survey}. 

%To retain the features most sufficient and get better performance for task, IB theory is extended to multi-view learning~\cite{xu2014large, lou2013multi}.

% The main contribution of these early researches was the introduction of information bottleneck theory into multi-view learning. 
%A trade-off between accuracy and complexity of multi-view model was found and the authors also revealed the robustness of multi-view learning and properties of generalization error bounds. 

The main challenge of applying the IB principle is that the mutual information terms are computationally intractable. In order to solve this problem, some researches have access to variational approximation to estimate mutual information such as~\cite{lee2021variational, wang2019deep, aguerri2019distributed,federici2020learning,song2021multicolor}. The main idea of variational approximation is to develop a variational bound on the sufficiency and redundancy trade-off. Specifically, the intractable distributions will be replaced by some known distributions like Gaussian distribution to calculate a lower bound of the original target. Then the optimal solution of the original objective is obtained by maximization of the lower bound. 
%\commentSY{what do you mean by ``complexity relevant trade-off"}.
%Taking deep multi-view learning as an example, the features are extracted by neural networks and the bound can be approximated by Markov sampling with stochastic gradient descent optimization. By the way, the variational approximation method often utilizes the reparameterization trick to improve the training efficiency.
%Although our work mainly focuses on supervised multi-view learning, the use of IB principle has increased to some extent for unsupervised multi-view learning. 
%To improve the generalization of an unsupervised multi-view representation model, the authors in
%The network based on IB principle balances the complementarity and consistency among multiple views and it has better performance on generalization and robustness. 
IB principle has also been applied for unsupervised multi-view learning. For example, \cite{wan2021multi} compress redundant information with a collaborative multi-view IB network. Meanwhile, \cite{federici2020learning} assumes that each view contains the same task relevant information, and therefore suggests maximizing the mutual information between latent representations extracted from each view to compress redundant information. These studies also maximize some sample-based differentiable mutual information lower bound instead of directly optimizing objective. As mentioned above, these methods have the drawback that it is difficult to guarantee the tightness of this approximation.

\section{Methodology}
\label{sec:methodology}

\begin{figure}[t]
\centering
\includegraphics[height=4.0cm]{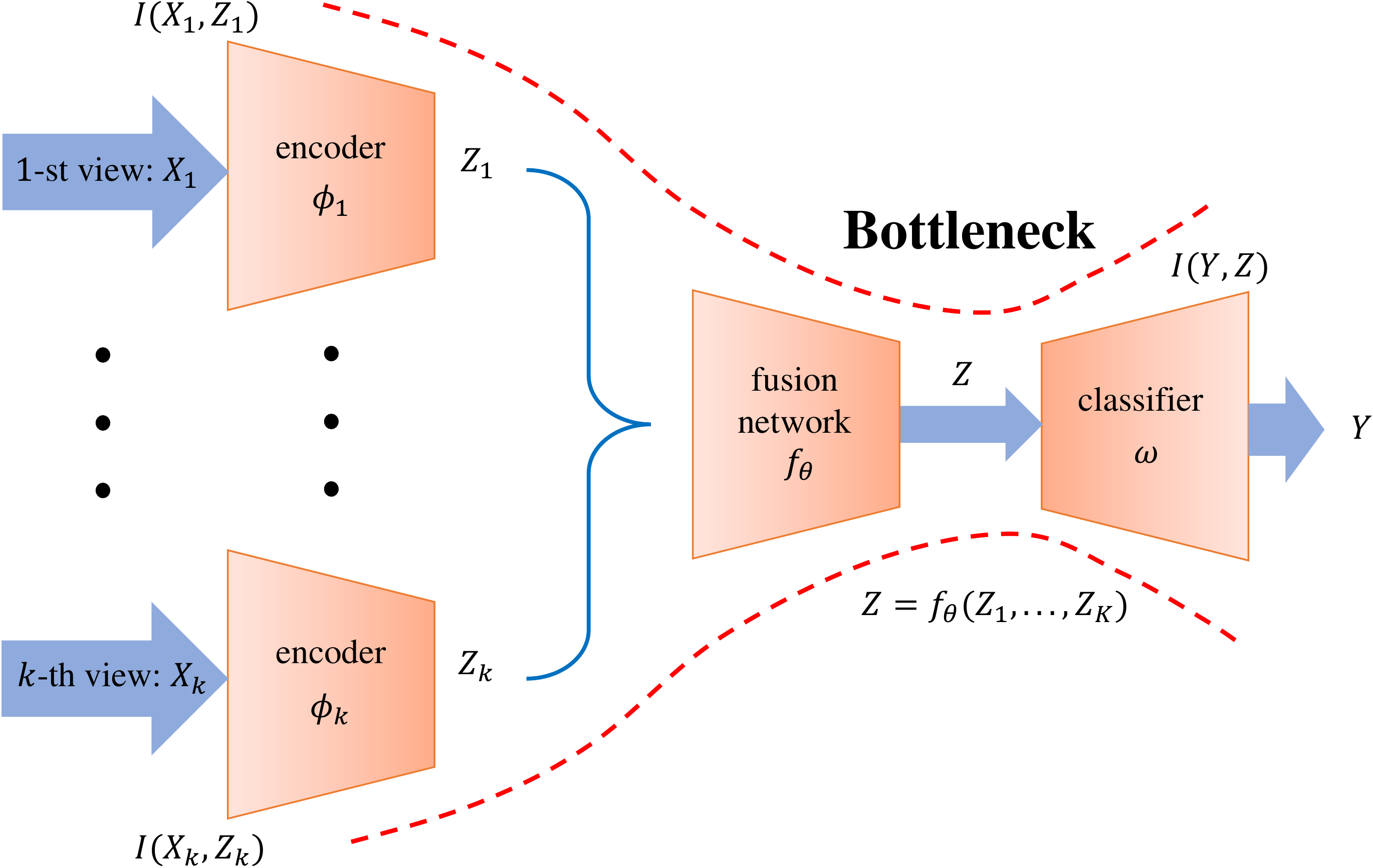}
\setlength{\parskip}{0pt}
\caption{An illustration of our proposed Multi-view matrix-Entropy based Information Bottleneck (MEIB). MEIB learns a robust joint representation $Z=f_\theta(Z_1,\cdots,Z_k)$ by striking a trade-off between $\max I(Y;Z)$ and $\min \sum_{i=1}^k I(X_i;Z_i)$.}
\label{fig_stru}
\end{figure}

The architecture of our MEIB is illustrated in Fig.~\ref{fig_stru}.
Specifically, we consider a general supervised multi-view learning scenario in which we aim to learn a joint yet robust representation $z$ given $k$ different views of features $(\mathbf{x}_1, \mathbf{x}_2, \cdots, \mathbf{x}_k)$ and their labels $y$. 
To this end, our MEIB consists of $k$ separate encoders ($\phi_1,\phi_2,\cdots,\phi_k$), a feature fusion network $f_\theta$, and a (nonlinear) classifier $w$. Each encoder maps view-specific data $\mathbf{x}_i$ ($1\leq i\leq k$) to a latent representation $\mathbf{z}_i$ to remove noise and redundant information contained in $\mathbf{x}_i$. The latent representations $\mathbf{z}_1,\mathbf{z}_2,\cdots,\mathbf{z}_k$ are then fused by a network $f_\theta$ to obtain a joint and compact representation $\mathbf{z}=f_\theta(\mathbf{z}_1,\mathbf{z}_2,\cdots,\mathbf{z}_k)$.

Note that, in multi-view unsupervised clustering, a common strategy to fuse view-specific latent representations is to take the form of $\mathbf{z}=\sum_{i=1}^k w_i \mathbf{z}_i$, s.t., $w_1+w_2+\cdots+w_k=1$ and $w_i\geq0$~\cite{trosten2021reconsidering}. Here, instead of using the na\"ive weighted linear combination, our fusion network $f_\theta$ is more general and absorbs the linear combiation as a special case.

% $Z_k$ is the latent features extracted from $v_k$ by $k$ encoders. 
%Then, the latent features are fused by $f_\theta$ for classification. 

Therefore, the overall objective of our MEIB can be expressed as:
\begin{equation}\footnotesize
\begin{aligned}
    \argmax_{\phi_1,\cdots,\phi_k,\theta,w} & I(Y;Z) - \sum_{i = 1}^k \beta _i I(X_i;Z_i), \\
    \text{s.t.} \quad & Z = f_\theta(Z_1,\cdots,Z_k),
\end{aligned}\label{eq:MEIB_objective}
\end{equation}
where $\beta_i$ refers to regularization parameter for the $i$-th view. 

% we maximize the mutual information between $Y$ and $Z$ which means the learned latent feature $Z$ contains information about label $Y$ as much as possible.

For the first term $\max I(Y;Z)$, it can be replaced with the risk associated with $Z$ to the prediction performance on $Y$ according to the cross-entropy loss $\texttt{CE}(Y;\hat{Y})$\footnote{The same strategy has also been used in the variational information bottleneck (VIB)~\cite{alemi2017deep}, the nonlinear information bottleneck (NIB)~\cite{kolchinsky2019nonlinear}, and the deep deterministic information bottleneck (DIB)~\cite{yu2021deep}.}~\cite{achille2018information,amjad2019learning}. Therefore, Eq.~(\ref{eq:MEIB_objective}) is equivalent to:
\begin{equation}\footnotesize
\begin{aligned}
    \argmin_{\phi_1,\cdots,\phi_k,\theta,w} & \texttt{CE}(Y,\hat{Y}) + \sum_{i = 1}^k \beta _i I(X_i;Z_i), \\
    \text{s.t.} \quad & Z = f_\theta(Z_1,\cdots,Z_k).
\end{aligned}\label{eq:MEIB_objective_CE}
\end{equation}

%To simplify the optimization, this term can be replaced by cross-entropy loss\footnote{The same trick used in deep IB \cite{wang2019deep} and VIB \cite{alemi2017deep}} which measures the dissimilarity between $\hat{Y}$ and $Y$. 

In this sense, the main challenge in optimizing Eq.~(\ref{eq:MEIB_objective}) or Eq.~(\ref{eq:MEIB_objective_CE}) is that the exact computation of the compression term $I(X_i;Z_i)$ is almost impossible or intractable due to the high dimensionality of the data.

In this work, we address this issue by simply making use of the matrix-based R{\'e}nyi's $\alpha$-order entropy functional to estimate $I(X;Z)$ in each view.
Specifically, given $N$ pairs of samples from a mini-batch of size $N$ in the $i$-th view, i.e., $\{\mathbf{x}^m,\mathbf{z}^m\}_{m=1}^{N}$, each $\mathbf{x}^m$ be an input sample and each $\mathbf{z}^m$ denotes its latent representation by encoder $\phi$\footnote{For simplicity, we remove the subscript $i$ of the view index in the remaining of this section.}. We can view both $\mathbf{x}$ and $\mathbf{z}$ as random vectors.
According to~\cite{giraldo2014measures}, the entropy of $\mathbf{x}$ can be defined over the eigenspectrum of a (normalized) Gram matrix $K_{\mathbf{x}}\in \mathbb{R}^{N\times N}$ ($K_{\mathbf{x}}(m,n)=\kappa(\mathbf{x}^{m}, \mathbf{x}^{n})$ and $\kappa$ is a Gaussian kernel) as:
\begin{equation}\label{eq:Renyi_entropy}\footnotesize
H_{\alpha}(A_\mathbf{x})=\frac{1}{1-\alpha}\log_2 \left(\tr (A_{\mathbf{x}}^{\alpha})\right)=\frac{1}{1-\alpha}\log_{2}\left(\sum_{m=1}^{N}\lambda _{m}(A_\mathbf{x})^{\alpha}\right),
\end{equation}
where $\alpha\in (0,1)\cup(1,\infty)$. $A_\mathbf{x}$ is the normalized version of $K_\mathbf{x}$, i.e., $A_\mathbf{x}=K_\mathbf{x}/\tr(K_\mathbf{x})$. $\lambda_{m}(A_\mathbf{x})$ denotes the $m$-th eigenvalue of $A_\mathbf{x}$.

The entropy of $\mathbf{z}$ can be measured similarly on the eigenspectrum of another (normalized) Gram matrix $A_\mathbf{z}\in \mathbb{R}^{N\times N}$.

Further, the joint entropy for $\mathbf{x}$ and $\mathbf{z}$ can be defined as:
\begin{equation}\label{eq:joint_entropy}\footnotesize
    H_\alpha (A_\mathbf{x},A_\mathbf{z}) = H_{\alpha}\left(\frac{A_\mathbf{x} \circ A_\mathbf{z}}{\tr (A_\mathbf{x} \circ A_\mathbf{z})}\right),
\end{equation}
where $\circ$ denotes Hadamard (or element-wise) product.

Given Eqs.~(\ref{eq:Renyi_entropy}) and (\ref{eq:joint_entropy}), the matrix-based R{\'e}nyi's $\alpha$-order mutual information $I_{\alpha}(X; Z)$ in analogy of Shannon's mutual information is given by:
\begin{equation}\label{eq:Renyi_MI}\small
I_{\alpha}(X;Z)=H_{\alpha}(A_\mathbf{x})+H_{\alpha}(A_\mathbf{z})-H_{\alpha}(A_\mathbf{x},A_\mathbf{z}).
\end{equation}

% \begin{comment}
% At the same time, we directly employ matrix-based Rényi's $\alpha$-order entropy to estimate the second term ${I}(V_i, Z_i)$ and make sure that the latent feature $Z$ contains less noise and extraneous information from $V_i$. Specifically, the matrix-based Rényi's $\alpha$-order mutual information $I_{\alpha}(V_{i},Z_{i})$ is given by~\cite{yu2021deep}:
% \begin{equation}
%     I_{\alpha}(V_{i},Z_{i})=H_{\alpha}(V_{i})+H_{\alpha}(Z_{i})-H_{\alpha}(V_{i}, Z_{i})
% \end{equation}
% where $\alpha  \in (0,1) \cap (1,\infty )$, and the $H_{\alpha}(V_{i})$ can be expressed by:
% \begin{equation}
%     {H_\alpha }({V_i}) = \frac{1}{{1 - \alpha }}{\log _2}(\tr({V_i}^\alpha )) = \frac{1}{{1 - \alpha }}{\log _2}(\sum\limits_{j = 1}^n {{{({\lambda _j}({V_i}))}^\alpha }} )
% \end{equation}
% and ${H_\alpha }({Z_i})$ is the same way, where ${\lambda _j}({V_i})$ denotes the $j$-th eigenvalue of $V_i$, $V_i$ is the normalized Gram matrices obtained by radial basis function kernel. $H_{\alpha}(V_{i}, Z_{i})$ is: 
% \begin{equation}
%     {H_\alpha }({V_i},{Z_i}) = {H_\alpha }(\frac{{{V_i} \circ {Z_i}}}{{tr({V_i} \circ {Z_i})}})
% \end{equation}
% where $\circ$ denotes Hadamard product.
% \end{comment}

The differentiability of matrix-based R{\'e}nyi's $\alpha$-order entropy functional has been derived in~\cite{yu2021measuring}. In practice, the automatic eigenvalue decomposition is embedded in mainstream deep learning APIs like TensorFlow and PyTorch. 

Compared with variational approximation or adversarial training that requires an additional auxiliary network (e.g., the mutual information neural estimator or MINE~\cite{belghazi2018mutual}) to approximate a lower bound of mutual information values, Eq.~(\ref{eq:Renyi_MI}) is mathematically well defined and computational efficient, especially for large networks. 
Moreover, there are only two hyper-parameters here: the order $\alpha$ and the kernel width $\sigma$. Throughout this work, we set $\alpha=1.01$ to approximate the Shannon entropy. For $\sigma$, we evaluate the $k$ ($k=10$) nearest distances of each sample and take the mean. We choose $\sigma$ as the average of mean values for all samples.

\section{EXPERIMENTS}
\label{sec:experiments}
In this section, we compare the performance of our MEIB with several popular multi-view learning methods on both synthetic and real-word datasets. The selected competitors include linear~CCA \cite{chaudhuri2009multi}, DCCA~\cite{andrew2013deep}, DCCAE~\cite{wang2015deep}, DNN, and deep IB~\cite{wang2019deep}. Here, DNN refers to a network shown in Fig.~\ref{fig_stru} trained by cross-entropy loss. We repeat the experiment 5 times to obtain average classification errors.  
%Specifically, we evaluate the performances on classification in terms of eliminating noise and ability of removing redundant features. We denote the proposed method as MEIB.
\subsection{Synthetic data}
\label{ssec:synthetic data}
 The method for synthesizing data is the same as in~\cite{wang2019deep}. We sample $s$ points per class from $N(0.5e, I)$ or $N(-0.5e, I)$ to form latent representation $Z$. To distort the classification, we concatenate the extra features $ext\_Z$ to the latent representation and synthesize new features $D$, i.e., $D = [Z, ext\_Z]$. Here the extra features for view $v_1$ contain $2s/3$ samples from $N(e, I)$ and $4s/3$ samples from $N(-e, I)$. The extra features for view $v_2$ contain $s/3$ samples from $N(e, I)$ and $5s/3$ samples from $N(-e, I)$. Then the nonlinear transformation of $D$ is accompanied by adding noise, which is sampled from $N(0, t*I)$, where $t$ denotes the noise level. Thus, the generation method of each view can be expressed as $X_i=f(D)+noise$, $i=\{1,2\}$. Of all the synthetic datasets, $80\%$ are used for training and $20\%$ for testing. 

\subsubsection{Robustness to Noise}
\label{sssec:Results about eliminating noise performance}
To evaluate the performances of all competing methods on their robustness to noise, we change the noise level $t$ which is calculated as $t=a\times \max(abs(v_{i}))$, where $a$ is set to be $[0.2 0.4 0.6 0.8 1.0 1.2]$. The dimension of extra features and latent representation are $5$ and $20$, respectively. The sample size $s$ per class is set to be $500$. So view $v_1$ can be expressed as  $v_1=tanh({tanh({Z;ext\_Z}))+0.1}$ and view $v_2$ can be expressed as $sigmoid({Z;ext\_Z})-0.5$. $\beta_1$ and $\beta_2$ are tuned in $[1e-5, 5e-5, 1e-4, 5e-4, 1e-3, 5e-3, 1e-2]$. For a fair comparison, in DNN, deep IB and our MEIB, the encoder for each view is a MLP with fully connected layers of the form $512 -512 -512$ for view $v_1$ and a single hidden layer $512$ for view $v_{2}$. The fusion layer to classification layer is the form of $256-10$ and all of the activate function is $ReLU$.
\begin{figure}[htb]
\centering
\includegraphics[height=2.8cm]{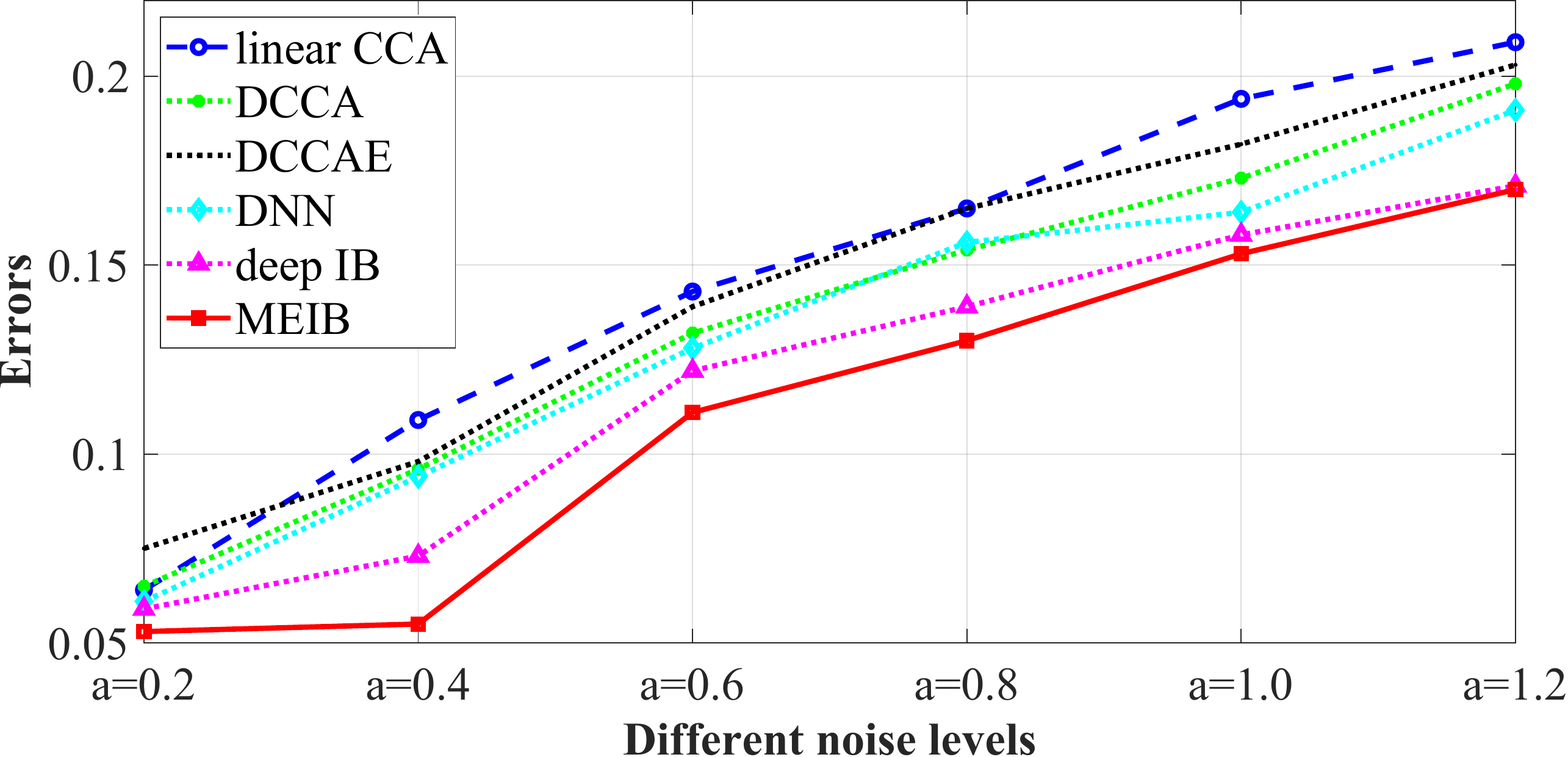}
\caption{``Classification error v.s. noise level" for all methods.}
\label{fig_noise}
\end{figure}
The experimental results of noise robustness are shown in Fig. \ref{fig_noise}. We can see that the performance of all competing methods decreases with the increase of noise level $a$. However, our MEIB and deep IB are consistently better than others, which indicates the advantage of bottleneck regularization, i.e., $I(X;Z)$, in suppressing noise. 

%Not surprisingly, the information bottleneck regularization, i.e., $I(X;Z)$, enables the removal 

%At the same time, the performance of supervised methods such as DNN is better than that of unsupervised methods such as linear CCA, which is also a reasonable result. More importantly, the information bottleneck principle appears to work well in this experiment.

\subsubsection{Robustness to Redundant Dimensions}
\label{ssec:Ability of removing redundant features}
\begin{figure}[htb]
    \centering
    \includegraphics[height=3.5cm]{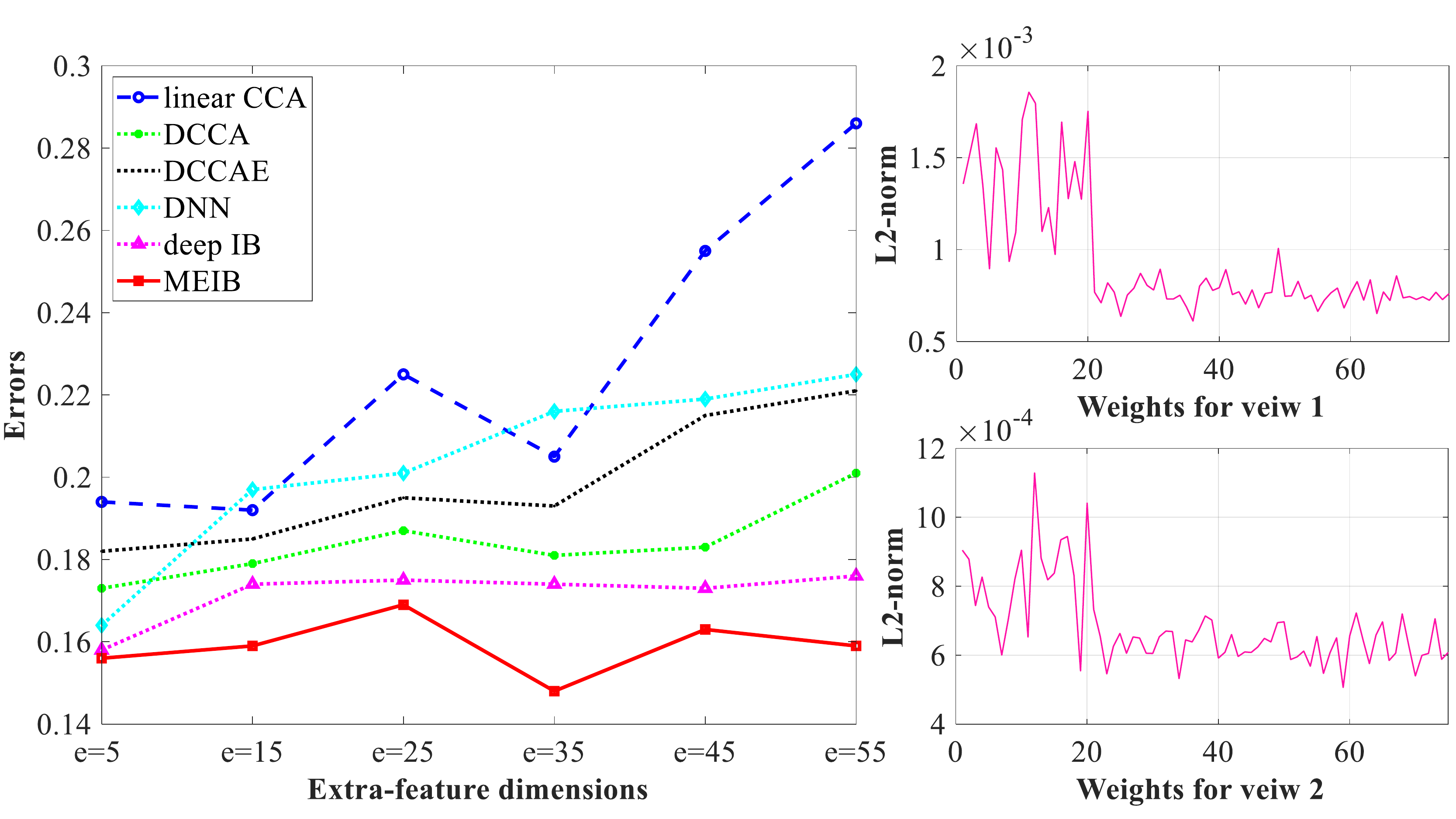}%width=7.5cm, 
    \caption{Robustness to redundant dimensions}
    \label{fig_exf}
\end{figure}

%In this section, we focus on the proposed method of removing redundant features ability. 
We further test the robustness of all competing methods to redundant dimensions. To this end, we change the extra features' dimension from $5$ to $55$, with an interval of $10$. All other settings are the same as in Section~\ref{sssec:Results about eliminating noise performance}, except that the noise factor $a$ is fixed to be $1$. The experimental results are shown in Fig.~\ref{fig_exf} (left). Again, our MEIB and deep IB are superior to other methods. We also plot the $\ell_2$ norm of the learned weights (by MEIB) associated with input dimensions $1$-$20$ (correspond to informative features) and dimensions $21$-$75$ (correspond to additional redundant features). As can be seen in Fig.~\ref{fig_exf} (right), our MEIB is able to identify redundant features dimensions and put small network weights therein.
%and there are a couple of things to be highlighted: First, in multi view learning, the methods based on information bottleneck have natural advantages in drop out redundant features. Compared with linear CCA, DCCA, DCCAE and DNN, the methods with information bottleneck such as deep IB and our proposed method can ensure stable performance advantages in the case of increasing redundancy features. This proves that information bottleneck itself is an effective and significant principle for removing redundant features.
% \footnote{We sampled 5000 data for training and 1000 for testing since some baselines time-consuming.}
% \footnote{For the same reason, we also sampled 5000 data for training and 1000 for testing.}
\subsection{Real-world dataset}
\label{Real-world dataset}
We evaluate our proposed method on 3 real-world multi-view datasets. \textbf{MNIST} has $2$ views. The first view is obtained by rotating raw images with an angle of $\theta$ following a uniform distribution in the range $[-\pi/4,\pi/4]$; the second view is obtained by adding uniformly distributed background noise.  \textbf{BBC}\footnote{\url{http://mlg.ucd.ie/datasets/bbc.html}} consists of $2,225$ documents from BBC news in five topical areas. The first view has $4,659$ features and the second view has $4,633$ features. \textbf{Wisconsin X-Ray MicroBeam  (XRMB)}~\cite{wang2015unsupervised} consists of $2$ views. The first view has $273$ features and the second view has $112$ features.
The performance of different methods is shown in Table~\ref{tab:real_word data}, which further indicates the superiority of our MEIB.
\begin{table}[htb]\small
  \centering
   \caption{Average classification error in real-world datasets}
  \label{tab:real_word data}
  \begin{tabular}{@{}cccc@{}}                \toprule 
   Methods      & MNIST            & BBC              & XRMB \\ \midrule
   linear CCA   & 0.423$\pm$0.002  &0.168$\pm$0.002  & 0.461$\pm$0.007 \\
   DCCA         & 0.255$\pm$0.002  &0.346$\pm$0.012  & 0.379$\pm$0.008 \\
   DCCAE        & 0.374$\pm$0.004  &0.279$\pm$0.002  & 0.383$\pm$0.001 \\ 
   DNN          & 0.195$\pm$0.018  &0.079$\pm$0.011  & 0.222$\pm$0.010 \\ 
   deep IB      & \underline{0.194$\pm$0.022 } &\underline{0.077$\pm$0.018 } & \underline{0.217$\pm$0.008} \\ 
   MEIB        & \textbf{0.173$\pm$0.022 }&\textbf{0.076$\pm$0.023 }  &\textbf{ 0.181$\pm$0.010 }\\\bottomrule
  \end{tabular} 
\end{table} 

% \commentSY{what is quantitative metrics, MSE or?Average classification error} 
\section{conclusion and future work}
\label{sec:refs}
%We have proposed a novel multi-view learning method with matrix-based Rényi's $\alpha$-order entropy and the empirical results show excellent performance in removing redundant features and eliminating the influence of noise. 

We developed the Multi-view matrix-Entropy based Information Bottleneck (MEIB) for multi-view learning. Using the recently proposed matrix-based R{\'e}nyi's $\alpha$-order entropy functional, MEIB can be optimized directly by SGD or ADAM, without variational approximation or adversarial training. Empirical results show that our MEIB outperforms other competing methods in terms of robustness to noise and redundant information contained in source data.

\begin{figure}[htb]
    \centering
    \includegraphics[width=6.0cm]{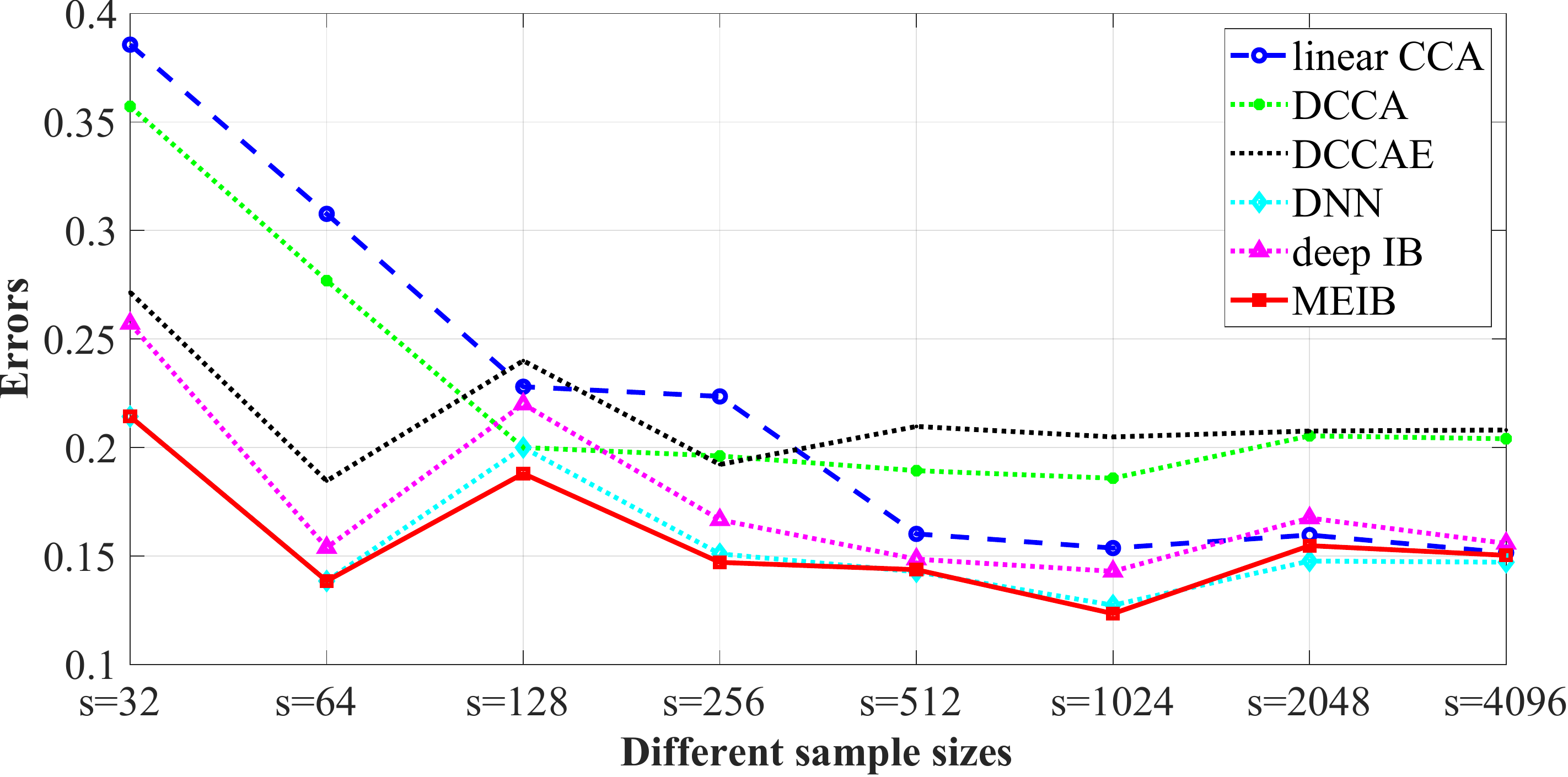}
    \caption{Classification error (on synthetic data) of all competing methods with respect to sample sizes $s$ in each view.}
    \label{fig_samp}
\end{figure}

Our study also raise a few insights (or open problems) concerning the general IB approach and its implementation:
\textbf{1) When is the matrix-based entropy necessary?}
As show in Fig.~\ref{fig_samp}, our advantages become weak with the increase of the number of samples in each view. This makes sense, because the increased sample size makes the variational approximation or the adversarial training with an additional network becomes stable. The lower bound of mutual information values also becomes tighter. 

% In future work, we hope to explore the performance with sample size increasing. We change the sample size in the range of $[32, 64, 128, 256, 512, 1024, 2048, 4096]$. All other settings are the same as setting of results 2, except that extra features dimension is set to be $5$.

\begin{figure}[htb]
    \centering
    \includegraphics[width=6.5cm]{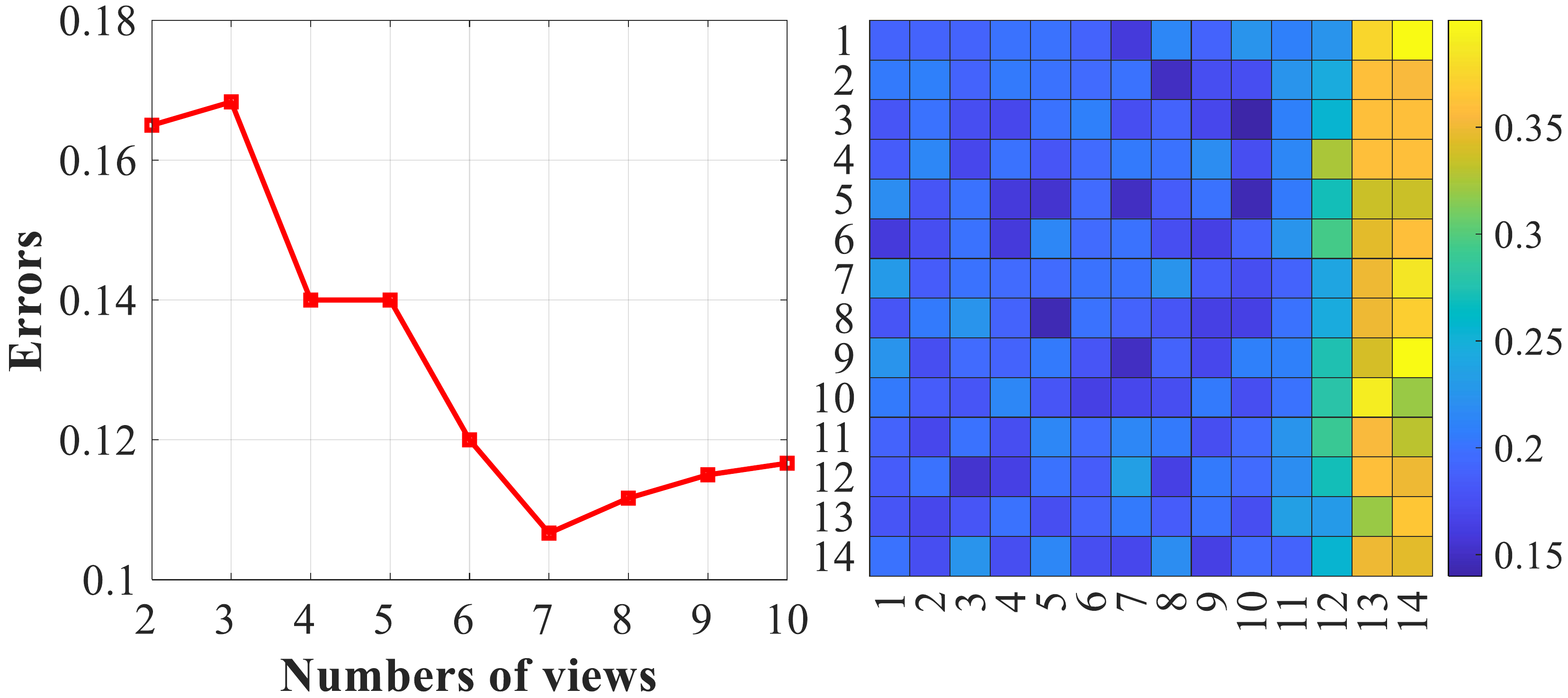}
    \caption{Left: error with respect to number of views for MEIB; Right: error with respect to different values of $\beta_1$ and $\beta_2$ (tuned in~$[0, 1e-5, 5e-5, 1e-4, 5e-4, 1e-3, 5e-3, 1e-2, 5e-2, 0.01, 0.05, 0.1, 1, 10]$).}
    \label{fig_10v}
\end{figure}

\noindent
\textbf{2) Practical concern for more than $2$ views?}
Extending existing multi-view learning methods to more than $2$ views is always a challenging problem~\cite{federici2020learning}. Although our framework can be simply extended to arbitrary number of views by just add a new regularization term $I(X_{\text{new}};Z_{\text{new}})$, it causes an additional hyper-parameter $\beta_{\text{new}}$ as well. Although we observed that the performance of our MEIB is stable for a suitable range of $\beta_i$ (Fig.~\ref{fig_10v} right), we also observed that increasing the number of views does not continually to improve classification performance (Fig.~\ref{fig_10v} left). This suggests that the IB regularization term $I(X;Z)$ alone is insufficient. One possible way is to take into consideration the ``synergistic" information~\cite{williams2010nonnegative} amongst all views in the future. 

% This is undesirable in case of multiple views. 

%In the future work, we hope to explore the performance with sample size increasing and analyze the performance of the information bottleneck against a large number of sample data.....

\vfill\pagebreak

% References should be produced using the bibtex program from suitable
% BiBTeX files (here: strings, refs, manuals). The IEEEbib.bst bibliography
% style file from IEEE produces unsorted bibliography list.
% -------------------------------------------------------------------------
\balance
\bibliographystyle{IEEEbib}
\bibliography{main}

\begin{thebibliography}{10}

\bibitem{lee2021variational}
Changhee Lee and Mihaela Schaar,
\newblock ``A variational information bottleneck approach to multi-omics data
  integration,''
\newblock in {\em AISTATS}. PMLR, 2021, pp. 1513--1521.

\bibitem{chaudhuri2009multi}
Kamalika Chaudhuri, Sham~M Kakade, Karen Livescu, and Karthik Sridharan,
\newblock ``Multi-view clustering via canonical correlation analysis,''
\newblock in {\em ICML}, 2009, pp. 129--136.

\bibitem{arora2012kernel}
Raman Arora and Karen Livescu,
\newblock ``Kernel cca for multi-view learning of acoustic features using
  articulatory measurements,''
\newblock in {\em SMLSLP}, 2012.

\bibitem{yan2021deep}
Xiaoqiang Yan, Shizhe Hu, Yiqiao Mao, Yangdong Ye, and Hui Yu,
\newblock ``Deep multi-view learning methods: a review,''
\newblock {\em Neurocomputing}, 2021.

\bibitem{andrew2013deep}
Galen Andrew, Raman Arora, Jeff Bilmes, and Karen Livescu,
\newblock ``Deep canonical correlation analysis,''
\newblock in {\em ICML}. PMLR, 2013, pp. 1247--1255.

\bibitem{wang2015deep}
Weiran Wang, Raman Arora, Karen Livescu, and Jeff Bilmes,
\newblock ``On deep multi-view representation learning,''
\newblock in {\em ICML}. PMLR, 2015, pp. 1083--1092.

\bibitem{tishby1999information}
Naftali Tishby, Fernando~C Pereira, and William Bialek,
\newblock ``The information bottleneck method,''
\newblock in {\em Allerton}, 1999, pp. 368--377.

\bibitem{xu2014large}
Chang Xu, Dacheng Tao, and Chao Xu,
\newblock ``Large-margin multi-view information bottleneck,''
\newblock {\em IEEE TPAMI}, vol. 36, no. 8, pp. 1559--1572, 2014.

\bibitem{lou2013multi}
Zhengzheng Lou, Yangdong Ye, and Xiaoqiang Yan,
\newblock ``The multi-feature information bottleneck with application to
  unsupervised image categorization,''
\newblock in {\em IJCAI}, 2013.

\bibitem{wang2019deep}
Qi~Wang, Claire Boudreau, Qixing Luo, Pang-Ning Tan, and Jiayu Zhou,
\newblock ``Deep multi-view information bottleneck,''
\newblock in {\em SDM}. SIAM, 2019, pp. 37--45.

\bibitem{aguerri2019distributed}
Inaki~Estella Aguerri and Abdellatif Zaidi,
\newblock ``Distributed variational representation learning,''
\newblock {\em IEEE TPAMI}, vol. 43, no. 1, pp. 120--138, 2019.

\bibitem{federici2020learning}
M~Federici, A~Dutta, P~Forr{\'e}, N~Kushmann, and Z~Akata,
\newblock ``Learning robust representations via multi-view information
  bottleneck,''
\newblock in {\em ICLR}, 2020.

\bibitem{song2021multicolor}
Jingqi Song et~al.,
\newblock ``Multicolor image classification using the multimodal information
  bottleneck network (mmib-net) for detecting diabetic retinopathy,''
\newblock {\em Optics Express}, vol. 29, no. 14, pp. 22732--22748, 2021.

\bibitem{giraldo2014measures}
Luis~Gonzalo Sanchez~Giraldo, Murali Rao, and Jose~C Principe,
\newblock ``Measures of entropy from data using infinitely divisible kernels,''
\newblock {\em IEEE TIT}, vol. 61, no. 1, pp. 535--548, 2014.

\bibitem{gilad2003information}
Ran Gilad-Bachrach, Amir Navot, and Naftali Tishby,
\newblock ``An information theoretic tradeoff between complexity and
  accuracy,''
\newblock in {\em Learning Theory and Kernel Machines}, pp. 595--609. Springer,
  2003.

\bibitem{sun2013survey}
Shiliang Sun,
\newblock ``A survey of multi-view machine learning,''
\newblock {\em NCAA}, vol. 23, no. 7, pp. 2031--2038, 2013.

\bibitem{wan2021multi}
Zhibin Wan, Changqing Zhang, Pengfei Zhu, and Qinghua Hu,
\newblock ``Multi-view information-bottleneck representation learning,''
\newblock in {\em AAAI}, 2021, pp. 10085--10092.

\bibitem{trosten2021reconsidering}
Daniel~J Trosten, Sigurd Lokse, Robert Jenssen, and Michael Kampffmeyer,
\newblock ``Reconsidering representation alignment for multi-view clustering,''
\newblock in {\em CVPR}, 2021, pp. 1255--1265.

\bibitem{alemi2017deep}
Alexander~A Alemi, Ian Fischer, Joshua~V Dillon, and Kevin Murphy,
\newblock ``Deep variational information bottleneck,''
\newblock in {\em ICLR}, 2017.

\bibitem{kolchinsky2019nonlinear}
Artemy Kolchinsky, Brendan~D Tracey, and David~H Wolpert,
\newblock ``Nonlinear information bottleneck,''
\newblock {\em Entropy}, vol. 21, no. 12, pp. 1181, 2019.

\bibitem{yu2021deep}
Xi~Yu, Shujian Yu, and Jos{\'e}~C Pr{\'\i}ncipe,
\newblock ``Deep deterministic information bottleneck with matrix-based entropy
  functional,''
\newblock in {\em ICASSP}. IEEE, 2021, pp. 3160--3164.

\bibitem{achille2018information}
Alessandro Achille and Stefano Soatto,
\newblock ``Information dropout: Learning optimal representations through noisy
  computation,''
\newblock {\em IEEE TPAMI}, vol. 40, no. 12, pp. 2897--2905, 2018.

\bibitem{amjad2019learning}
Rana~Ali Amjad and Bernhard~C Geiger,
\newblock ``Learning representations for neural network-based classification
  using the information bottleneck principle,''
\newblock {\em IEEE TPAMI}, vol. 42, no. 9, pp. 2225--2239, 2019.

\bibitem{yu2021measuring}
Shujian Yu, Francesco Alesiani, Xi~Yu, Robert Jenssen, and Jose Principe,
\newblock ``Measuring dependence with matrix-based entropy functional,''
\newblock in {\em AAAI}, 2021, pp. 10781--10789.

\bibitem{belghazi2018mutual}
Mohamed~Ishmael Belghazi, Aristide Baratin, Sai Rajeshwar, Sherjil Ozair,
  Yoshua Bengio, Aaron Courville, and Devon Hjelm,
\newblock ``Mutual information neural estimation,''
\newblock in {\em ICML}. PMLR, 2018, pp. 531--540.

\bibitem{wang2015unsupervised}
Weiran Wang, Raman Arora, Karen Livescu, and Jeff~A Bilmes,
\newblock ``Unsupervised learning of acoustic features via deep canonical
  correlation analysis,''
\newblock in {\em ICASSP}. IEEE, 2015, pp. 4590--4594.

\bibitem{williams2010nonnegative}
Paul~L Williams and Randall~D Beer,
\newblock ``Nonnegative decomposition of multivariate information,''
\newblock {\em arXiv preprint arXiv:1004.2515}, 2010.

\end{thebibliography}
\end{document}